\documentclass[runningheads]{llncs}
\usepackage{xcolor}
\usepackage{graphicx}
\usepackage{booktabs,multicol,multirow}

\begin{document}
\title{Right for the Wrong Reason: Can Interpretable ML Techniques Detect Spurious Correlations?}
\titlerunning{Right for the Wrong Reason}

\author{Susu Sun\inst{1}\and
Lisa M. Koch\inst{2,3} \and
Christian F. Baumgartner\inst{1}}
% index{Sun, Susu} 
% index{Koch, Lisa M} 
% index{Baumgartner, Christian F} 

\authorrunning{Sun et al.}

\institute{Cluster of Excellence -- ML for Science, University of Tübingen, Germany \and
Hertie Institute for AI in Brain Health, University of Tübingen, Germany \and
Institute of Ophthalmic Research, University of Tübingen, Germany\\
\email{\{susu.sun,lisa.koch,christian.baumgartner\}@uni-tuebingen.de}}

\maketitle              % typeset the header of the contribution
\begin{abstract}
While deep neural network models offer unmatched classification performance, they are prone to learning spurious correlations in the data. Such dependencies on confounding information can be difficult to detect using performance metrics if the test data comes from the same distribution as the training data. Interpretable ML methods such as post-hoc explanations or inherently interpretable classifiers promise to identify faulty model reasoning. However, there is mixed evidence whether many of these techniques are actually able to do so. In this paper, we propose a rigorous evaluation strategy to assess an explanation technique's ability to correctly identify spurious correlations. Using this strategy, we evaluate five post-hoc explanation techniques and one inherently interpretable method for their ability to detect three types of artificially added confounders in a chest x-ray diagnosis task. We find that the post-hoc technique SHAP, as well as the inherently interpretable Attri-Net provide the best performance and can be used to reliably identify faulty model behavior. 


\end{abstract}

\section{Introduction}

Black-box neural network classifiers offer enormous potential for computer-aided diagnosis and prediction in medical imaging applications but, unfortunately, they also have a strong tendency to learn spurious correlations in the data~\cite{geirhos2020shortcut}. For the development and safe deployment of machine learning (ML) systems it is essential to understand what information the classifiers are basing their decisions on, such that reliance on spurious correlations may be identified. 

Spurious correlations arise when the training data are confounded by additional variables that are unrelated to the diagnostic information we want to predict. For instance, older patients in our training data may be more likely to present with a disease than younger patients. A classifier trained on this data may inadvertently learn to base its decision on image features related to age rather than pathology. Crucially, such faulty behavior cannot be identified using classification performance metrics such as area under the ROC curve (AUC) if the testing data contains the same confounding information as the training data, since the classifier predicts the \textit{right} thing, but for the \textit{wrong} reason. If undetected, however, such spurious correlations may lead to serious safety implications after deployment.

\begin{figure}
\includegraphics[width=\textwidth]{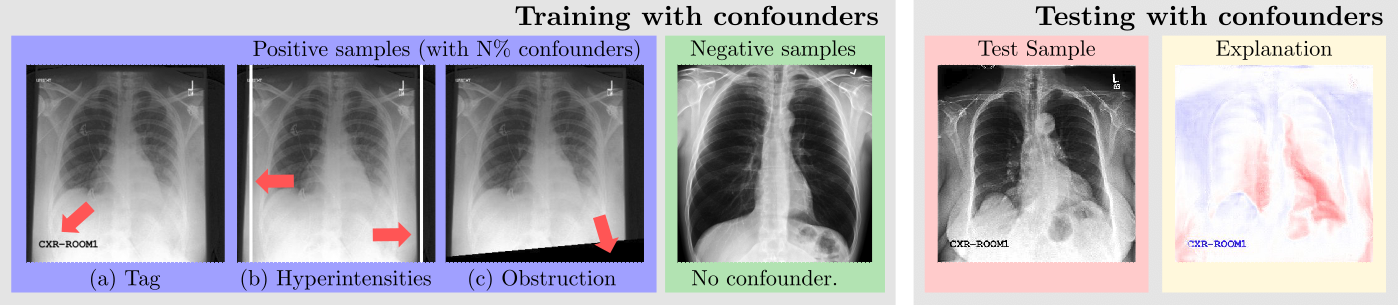}
\caption{\textbf{Overview.} We train classifiers on datasets with three types of artificially added confounders highlighted by arrows. We then evaluate the ability of explanation techniques to correctly identify reliance on these confounders (shown Attri-Net~\cite{sun2023inherently}).} \label{fig:overview}
\end{figure}

Interpretable ML approaches may be used as a powerful tool to detect spurious correlations during development or after deployment of an ML system.  
Currently, the most widely used explanation modality are \textit{visual} explanations, which highlight the pixels in the input image that are responsible for a particular decision. Common strategies include methods which leverage the gradient of the prediction with respect to the input image~\cite{springenberg2014striving,sundararajan2017axiomatic,smilkov2017smoothgrad,selvaraju2017grad,boreiko2022visual}, explain the predictions by counterfactually generating an image of the opposite class~\cite{cohen2021gifsplanation,samangouei2018explaingan,singla2019explanation,sun2023inherently}, interpret the feature map of the last layer before the classification~\cite{zhou2016learning,brendel2019approximating,djoumessi2023sparse}, or methods that build a local approximation of the decision function such as LIME~\cite{ribeiro2016should}, or SHAP~\cite{lundberg2017unified}. %We note that those categories are not necessarily mutually exclusive. 

The majority of visual explanation methods are \textit{post-hoc} techniques, meaning a heuristic is applied to any trained model (e.g. a ResNet~\cite{he2016deep}) to approximately understand the decision mechanism for a given data point. However, post-hoc techniques are by definition only approximations and many techniques have been found to suffer from serious limitations~\cite{white2019measurable,alvarez2018robustness,han2022explanation}. \textit{Inherently interpretable} techniques on the other hand build custom architectures that are designed to directly reveal the reasoning of the classifier to the user without the need for approximations. This class of methods does not suffer from the same limitations as post-hoc methods, and it has been argued that inherently interpretable approaches should be preferred in high-stakes applications such as medical image analysis~\cite{rudin2019stop}. For instance, if a classifier bases its decision on a spurious signal, an inherently interpretable classifier should by definition reveal this relationship. 

Inherently interpretable visual explanation approaches are much less widely explored than post-hoc techniques, but there has recently been an increased interest in the topic. Two recently proposed methods in this category are the attribution network (Attri-Net)~\cite{sun2023inherently}, and convolutional dynamic alignment networks (CoDA-Nets)~\cite{bohle2021convolutional}. Attri-Net first produces human-interpretable feature attribution maps for each disease category using a GAN-based counterfactual generator~\cite{sun2023inherently}. Then makes the final prediction with simple logistic regression classifiers based on those feature attribution maps.
CoDA-Nets express neural networks as input dependent linear transformation~\cite{bohle2021convolutional}. Both approaches produce explanations on the pixel level of the input images. 

\subsubsection{Related work on comparing explanation techniques}

A number of works have studied the quality of post-hoc explanation techniques. The vast majority of work focuses exclusively on gradient-based approaches  (e.g.~\cite{sixt2020explanations,arun2021assessing}). In their landmark study, Adebayo et al.~\cite{adebayo2018sanity} find that commonly used gradient-based explanation techniques do not pass some basic sanity checks. Arun et al.~\cite{arun2021assessing} extends this work to weakly supervised localisation in one of the few papers in this domain focusing on medical data. Both papers, however, do not consider other types of commonly used approaches such as counterfactual methods, or local function approximations such as LIME or SHAP. 

A small number of works specifically investigate explanations' sensitivity to spurious correlations. In closely related work to ours, Adebayo et al.~\cite{adebayo2020debugging} explore a large library of post-hoc explanation techniques including LIME and SHAP, for detecting spurious image backgrounds in a bird versus dog classification task and find that many techniques are in fact able to detect the spurious background. In subsequent work, the same authors explore the usefulness of four post-hoc gradient-based explanation methods for identifying spurious correlations in hand and knee radiographs~\cite{adebayo2022post} and come to the conclusion that the examined methods are ineffective at identifying spurious correlations. We note that prior work is inconclusive on the usefulness of explanation techniques for identifying spurious correlations. In particular, in the medical context it is still unclear if commonly used explanation techniques are suitable for the detection of spurious correlations. Moreover, there is, to our knowledge, no evidence for the supposition that inherently interpretable techniques are better suited for this task. 

\subsubsection{Contributions}

We present a rigorous evaluation of post-hoc explanations and inherently interpretable techniques for the identification of spurious correlations in a medical imaging task. Specifically, we focus on the task of diagnosing cardiomegaly from chest x-ray data with three types of synthetically generated spurious correlations (see Fig.~\ref{fig:overview}). To identify whether an explanation correctly identifies a model's reliance on spurious correlations, we propose two quantitative metrics which are highly reflective of our qualitative findings. In contrast to the majority of prior work we focus on a wide range of different explanation approaches including counterfactual techniques and local function approximations, as well as post-hoc techniques and an inherently interpretable approach. Our analysis yields actionable insights which will be useful for a wide audience of ML practitioners.

\section{A Framework for Evaluating Explanation Techniques}

In the following, we introduce our evaluation strategy and proposed evaluation metrics, the studied confounders, as well as the evaluated explanation techniques. The strategy and evaluation metrics are generic and can also be applied to different problems. The confounders are engineered to correspond to realistic image artifacts that can appear in chest x-ray imaging\footnote{Our code can be found under \url{github.com/ss-sun/right-for-the-wrong-reason}.}. 

\subsection{Evaluation Strategy}

We assume a setting in which the development data for a binary neural network based classifier contains an unknown spurious correlation with the target label. To quantitatively study this setting, we create training data with artificial spurious correlations by adding a confounding effect (e.g. a hospital tag) in a percentage of the cases with a positive label, where we vary the percentage $p \in \{0, 20, 50, 80, 100\}$. E.g., for $p=100\%$ all of the positive images in the training set will have an artificial confounder, and for $p=0\%$ there is no spurious signal. With increasing $p$ the reliance on a spurious signal becomes more likely. The images with a negative label remain untouched. 

In the evaluation, we consider a scenario in which the test data contain the same confounder type with the same proportion $p$ used in the respective trainings. In this case, we can not tell if a classifier relies on the confounded features from classification performance. Our aim, therefore, is to investigate whether explanation techniques can identify that the classifier predicts the \textit{right thing for the wrong reason}. 

We perform all experiments on chest x-ray images from the widely used CheXpert dataset~\cite{irvin2019chexpert}, where we focus on the binary classification task on disease cardiomegaly.
We divided our dataset into a training ($80\%$), validation  ($10\%$) test  ($10\%$) set.  

\subsection{Studied Confounders}

We study three types of confounders inspired by real-world artefacts. Firstly, we investigate a hospital tag placed in the lower left corner of the image (see Fig~\ref{fig:overview}a). Secondly, we add vertical lines of hyperintense signal that can be caused by foreign materials on the light path assembly (see Fig~\ref{fig:overview}b). Lastly, we consider an oblique occlusion of the image in the lower part of the image, which is an artefact that we observed for many images in the CheXpert dataset (see Fig~\ref{fig:overview}c). 

\subsection{Evaluation Metrics for Measuring Confounder Detection}

We propose two novel metrics which reflect an explanation's ability to correctly identify spurious correlations. \\ \vspace{-2mm}

\noindent \textbf{Confounder Sensitivity (CS)} 
Firstly, the explanations should be able to correctly attribute the confounder if classifier bases its decision on it. We assess this property by summing the number of true positive attributions divided by the total number of confounded pixels for each test image. We consider a pixel a true positive if it is part of the pixels affected by the confounder \textit{and} in the top 10\% attributed pixels according to a visual explanation. Thus the maximum sensitivity of 1 is obtained if all confounded pixels are in the top 10\% of the attributions. Note that we do not penalise attributions outside of the confounding label as those can still also be correct. To guarantee that we only evaluate on samples for which the prediction is actually influenced by the confounder, we only include images for which the prediction with and without the confounding label is of the opposite class. To reduce computation times we use a maximum of 100 samples for each evaluation. An optimal explanation methods should obtain a CS score of 0 if the data contains $p=0\%$ confounded data points, since in that case the spurious signal should not be attributed. For increasing $p$ the confounder sensitivity should increase, i.e. the explanation should reflect the classifiers increasing reliance on the confounder. \\ \vspace{-2mm}

\noindent \textbf{Sensitivity to prediction changes via explanation NCC}
Secondly, the explanations should not be invariant to changes in classifier prediction. That is, if the classifier's prediction for a specific image changes when adding or removing a confounder, then the explanations should also be different. We measure this property using the average normalised cross correlation (NCC) between explanations of test images when confounders were either present or absent.%
%, and expect a low NCC value for ideal explanation techniques for $p>0\%$. 
Again, we only evaluate on images for which the prediction changes when adding the confounder as in these cases, we know the classifier is relying on confounders, and we evaluate a maximum of 100 samples.  
An optimal explanation method should obtain a high NCC score if the training data contains $p=0\%$ confounded data points, since in that case the explanation with and without the confounder should be similar. 
For increasing $p$ the NCC score should decrease to reflect the classifiers increasing reliance on the confounder. 

\subsection{Evaluated Explanation Methods}

We evaluated five post-hoc techniques with representative examples from the approaches mentioned in the introduction: Guided Backpropgation~\cite{springenberg2014striving} and Grad-CAM~\cite{selvaraju2017grad} (gradient-based), Gifsplanation (counterfactual), and LIME~\cite{ribeiro2016should} and SHAP partition explainer \cite{lundberg2017unified} (local linear approximations). All post-hoc techniques were applied to a standard black-box ResNet50 model. We furthermore investigated the interpretable visual explanation method Attri-Net \cite{sun2023inherently}. We used the default parameters for all methods. We found CoDA-Nets \cite{bohle2021convolutional} required lengthy hyperparameter tuning for each type of experiment, and decided to exclude it in this paper.

%We also performed preliminary experiment with CoDA-Nets \cite{bohle2021convolutional}. However, we found that CoDA-Nets required lengthy hyperparameter tuning for each type of experiment. We thus decided to exclude CoDA-Nets from our results presented in this paper. 

\section{Results}

We first established the classifiers' performance in the presence of confounders, then compared all techniques in their ability to identify such confounders. \\

\begin{figure}
\includegraphics[width=\textwidth]{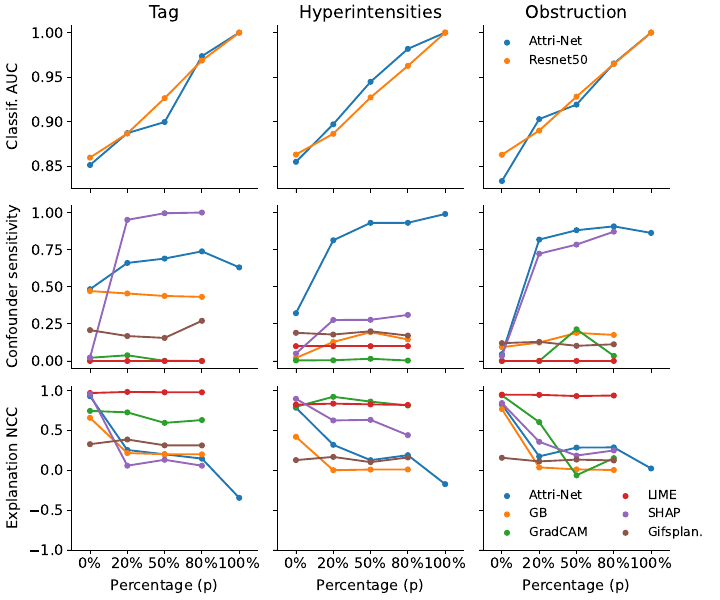}
\caption{\textbf{(top row)} Classification AUC of Attri-Net and Resnet50 on images containing hospital tags (left), hyperintensities (middle) or obstruction confounders (right column). The classifiers were trained with a varying proportion of confounders present in the positive examples in the training set (shown on the x-axes). \textbf{(bottom rows)} The explanation techniques' ability to identify confounders in terms of confounder sensitivity (middle row) and explanation NCC (bottom row, lower is better).} \label{fig:exp1}
\end{figure}

\noindent \textbf{Classification performance}
Both investigated classifiers, the ResNet50 and the inherently interpretable Attri-Net, performed similarly in terms of classification AUC (first row of Fig.~\ref{fig:exp1}). 
For all three confounders, classification AUC consistently increased with increasing contamination $p$ of the training dataset. This indicated that the classifiers increasingly relied on the spurious signal. For $p=100\%$ contamination, where the confounder was present on all positive training examples, both classifiers reached almost a perfect classification AUC of 1. 

\noindent \textbf{Explanations}
We analysed the explanations' ability to identify confounders by reporting confounder sensitivity (CS, middle row in Fig.~\ref{fig:exp1}) and explanation NCC (bottom row in Fig.~\ref{fig:exp1}). Out of the investigated methods Attri-Net and SHAP were closest to the ideal behaviour of high confounder sensitivity and low explanation NCC for $p>0\%$. We found that SHAP performed extremely well in detecting tag confounders, but struggled with hyperintensities confounders. This can be explained by the fact that the tag confounder is relatively small and thus is more likely to be completely covered by the superpixels in SHAP. Overall, the inherently interpretable Attri-Net technique achieved the best balance. In agreement with related literature we found that gradient-based explanation methods performed poorly. In particular, Guided Backpropagation displayed similar CS-scores no matter if the classifier relies on a spurious signal ($p>0\%$) or not ($p=0\%$). Note that some results for $p=100\%$ were missing because no data points fulfilled the criterion of the prediction being flipped with and without the confounders.

\begin{figure}
\includegraphics[width=\textwidth]{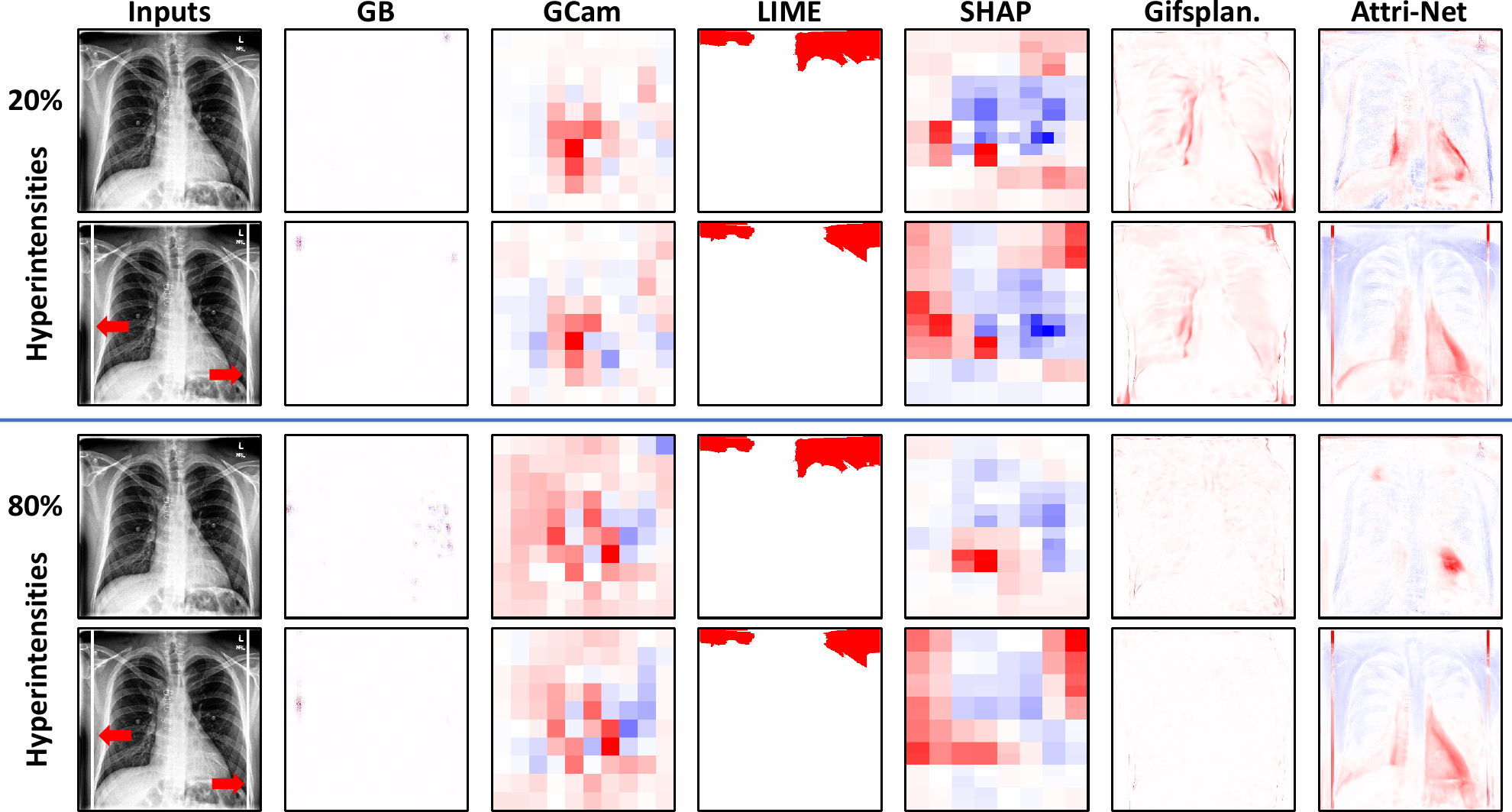}
\ \caption{Explanations for one example image with and without hyperintensities confounders. We show results for models trained on 20\% \textbf{(top rows)} and 80\% \textbf{(bottom rows)} confounded data points, respectively.} 
\label{fig:exps-hyperintensities-scales}
\end{figure}

\begin{figure}
\includegraphics[width=\textwidth]{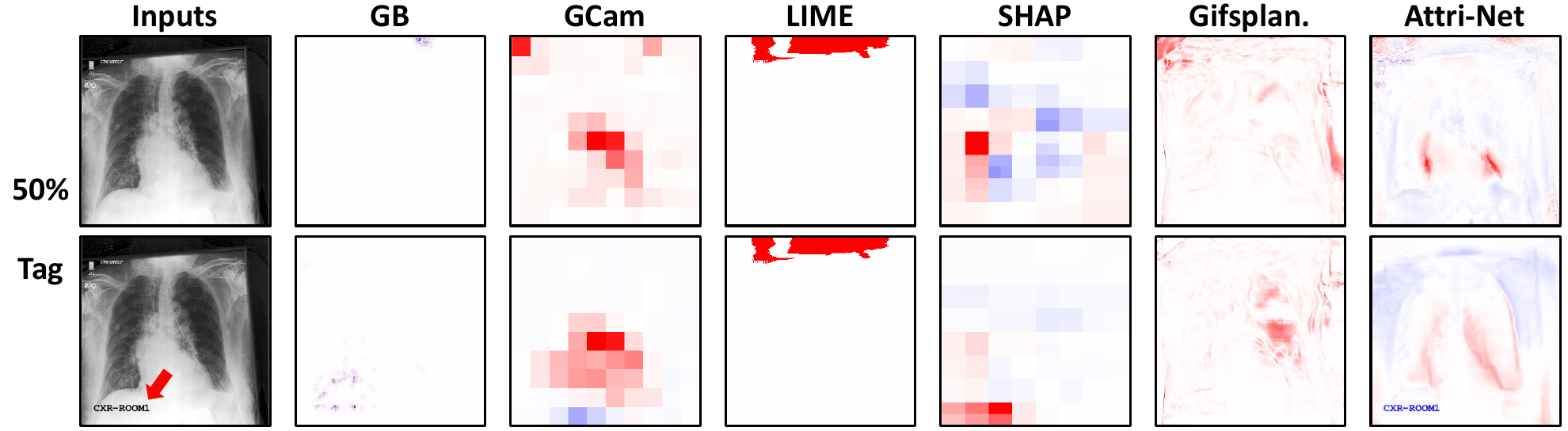}
\caption{Explanations for one example image with  \textbf{(top)} and without \textbf{(bottom)} a tag confounder  for models trained on 50\% confounded data points.} \label{fig:exps-tag}
\end{figure}

\begin{figure}
\includegraphics[width=\textwidth]{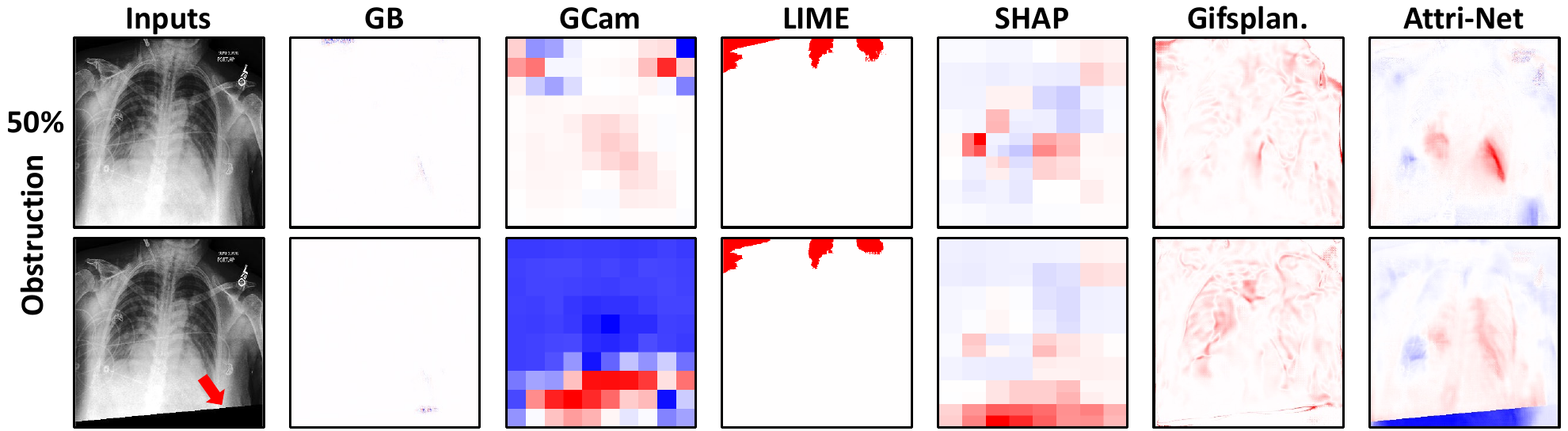}
\caption{Explanation for one example image  with \textbf{(top)} and without \textbf{(bottom)} an obstruction confounder for models trained on 50\% confounded data points.} \label{fig:exps-obstruction}
\end{figure}

Figs.~\ref{fig:exps-hyperintensities-scales}, \ref{fig:exps-tag} \& \ref{fig:exps-obstruction} contain examples explanations for the hyperintensity, tag, and edge confounder, respectively. Our qualitative analysis of the results confirms the quantitative findings, with SHAP and Attri-Net providing the most intuitive explanations. In particular, in the challenging hyperintensities scenario (see Fig.~\ref{fig:exps-hyperintensities-scales}) AttriNet was the only method able to highlight the confounders in a human-interpretable fashion. We note that in all examples when a confounder was present, SHAP tended to highlight only the confounder, while Attri-Net also highlighted features related to Cardiomegaly. This may reflect the different decision mechanisms of the ResNet50 and the Attri-Net. 

\section{Discussion}

In this paper, we proposed an evaluation strategy to assess the ability of visual explanations to correctly identify a classifier's reliance on a spurious signal. We specifically focused on the scenario where the classifier is predicting the right thing, but for the wrong reason, which is highly significant for the safe development of ML-basd diagnosis and prediction systems. Using this strategy, we assessed the performance of five post-hoc explanation techniques and one inherently interpretable technique with three realistic confounding signals. We found that the inherently interpretable Attri-Net technique, as well as the post-hoc SHAP technique performed the best, with Attri-Net yielding the most balanced performance. Both techniques are suitable for finding false reliance on a spurious signals. We also observed that the variation in the explanations' sparsity makes them perform differently in detecting spurious signals of different sizes and shapes. In agreement with prior work, we found that gradient based techniques performed less robustly in our experiments. 

From our experiments we draw two main conclusions. Firstly, practitioners looking to check for spurious correlations in a trained black-box model such as a ResNet should give preference to SHAP which provided the best performance out of the post-hoc techniques in our experiments. Secondly, an inherently interpretable technique, namely Attri-Net, performed the best in our experiments providing evidence to the supposition by Rudin et al.~\cite{rudin2019stop} that inherently interpretable techniques may provide a fruitful avenue for future work. 

A major limitation of our study is the limited number of techniques we examined. Thus a primary focus of future work will be to scale our experiments to a wider range of techniques. Future work will also focus on human-in-the-loop experiments, as we believe, this will be the ultimate assessment of the usefulness of different explanation techniques. 

\section*{Acknowledgements}
Funded by the Deutsche Forschungsgemeinschaft (DFG, German Research Foundation) under Germany’s Excellence Strategy – EXC number 2064/1 – Project number 390727645. The authors acknowledge support of the Carl Zeiss Foundation in the project ``Certification and Foundations of Safe Machine Learning Systems in Healthcare" and the Hertie Foundation.
The authors thank the International Max Planck Research School for Intelligent Systems (IMPRS-IS) for supporting Susu Sun, Lisa M. Koch, and Christian F. Baumgartner.

\bibliographystyle{splncs04}
\bibliography{references}

\title{Supplementary Materials}
\author{}
\institute{}

\maketitle    % typeset the header of the contribution

\begin{figure}
\includegraphics[width=\textwidth, scale=0.8]{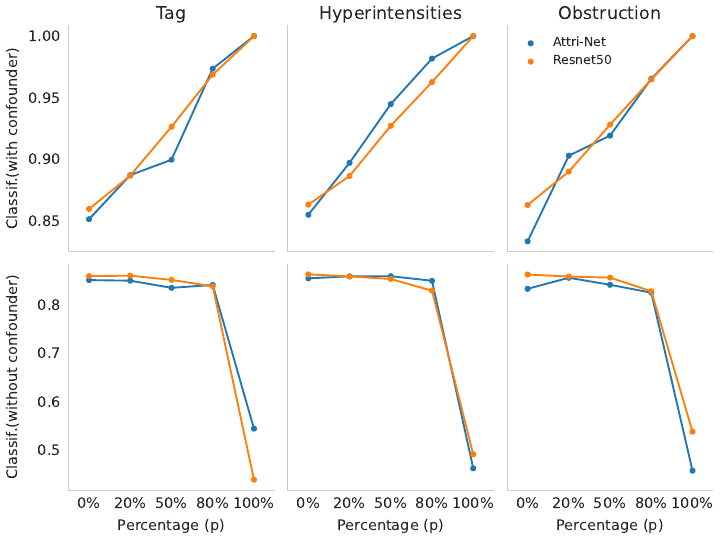}
\caption{\textbf{(top row)} Classification performance of Attri-Net and Resnet50 on a test set with the same proportion of confounders as the train set (same as top row of Fig. 2 in main manuscript). \textbf{(bottom row)} Classification performance of Attri-Net and Resnet50 on test set \textit{without} confounders. This demonstrates the models' reliance on the confounders and exemplifies potential risks associated with the deployment of such a model.} \label{fig:exp1_supple}
\vspace{-0.5cm}
\end{figure}

\begin{table}
\centering
\vspace{-0.3cm}
\caption{Correlation of confounder sensitivity and explanation NCC with contamination $p$.}\label{tab1}
\begin{tabular}{|c|c|c|c|c|c|c|}
\hline
Methods &  Attri-Net & LIME & GB & SHAP & GradCAM & Gifsplan.\\
\hline
Corr (confounder sensitivity, $p$)&  0.65 & 0 & 0.14 & 0.58 & 0.13 & 0.09\\
\hline
Corr (explanation NCC, $p$) &  -0.83 & -0.03 & -0.66 & -0.65 & -0.44 & -0.06\\
\hline
\end{tabular}
\vspace{-0.5cm}
\end{table}

\begin{figure}
\includegraphics[width=\textwidth]{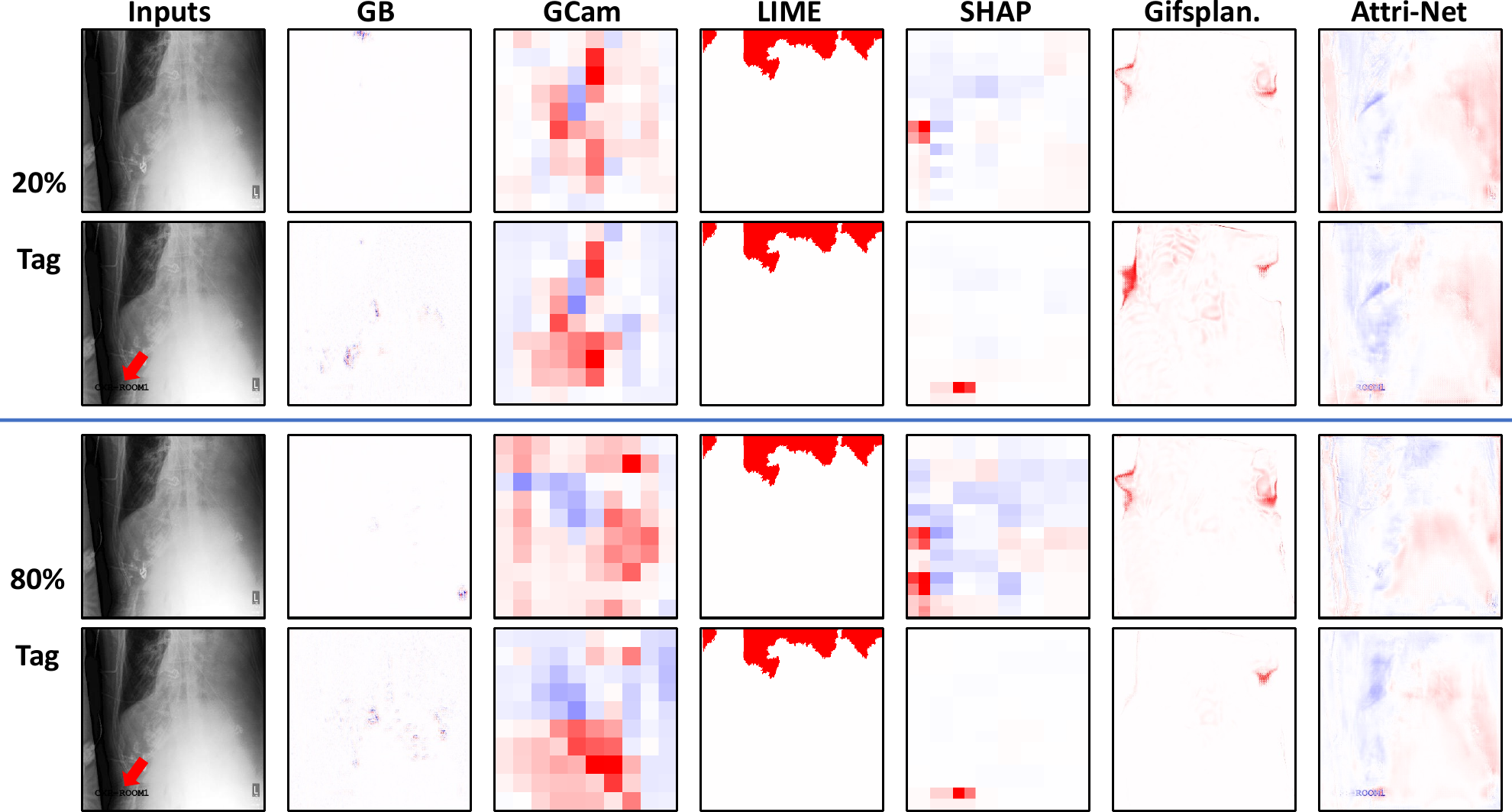}
\ \caption{Explanations for one example image with and without \textit{tag confounders}. We show results for models trained on 20\% \textbf{(top rows)} and 80\% \textbf{(bottom rows)} confounded data points, respectively.} 
\label{fig:exps-tag-scales}
\end{figure}

\begin{figure}
\includegraphics[width=\textwidth]{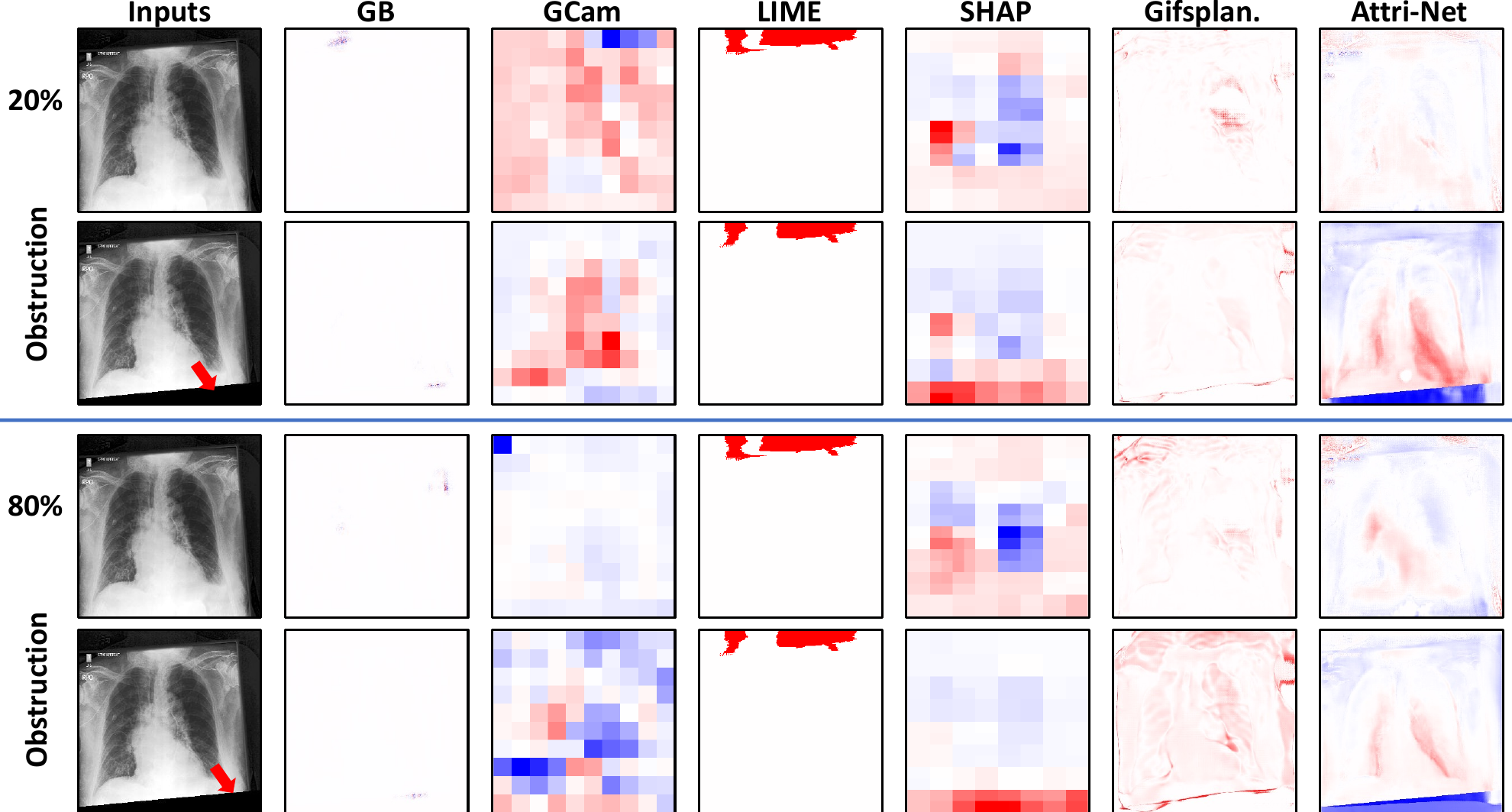}
\ \caption{Explanations for one example image with and without \textit{obstruction confounders}. We show results for models trained on 20\% \textbf{(top rows)} and 80\% \textbf{(bottom rows)} confounded data points, respectively.} 
\label{fig:exps-obstruction-scales}
\end{figure}

\end{document}